\documentclass[10pt,twocolumn,letterpaper]{article}

\usepackage[pagenumbers]{cvpr} % To force page numbers, e.g. for an arXiv version

\usepackage[dvipsnames]{xcolor}

\usepackage{graphicx}
\usepackage{amsmath}
\usepackage{amssymb}
\usepackage{booktabs}
\usepackage{caption}
\usepackage{colortbl}
\usepackage{makecell}
\usepackage{float}
\usepackage{paralist}
\usepackage{xcolor}
\usepackage{threeparttable}
\usepackage{algorithm}
\usepackage{algorithmic}
\usepackage{comment}

\newcommand{\tao}[1]{~\textcolor{red}{Tao:\ #1}}

\definecolor{cvprblue}{rgb}{0.21,0.49,0.74}
\usepackage[pagebackref,breaklinks,colorlinks,citecolor=cvprblue]{hyperref}
\usepackage{amsmath,amsfonts,bm}

\def\1{\bm{1}}

\def\vzero{{\bm{0}}}

\def\vf{{\bm{f}}}

\def\vs{{\bm{s}}}

\def\vw{{\bm{w}}}
\def\vx{{\bm{x}}}
\def\vy{{\bm{y}}}
\def\vz{{\bm{z}}}

\def\mI{{\bm{I}}}

\DeclareMathAlphabet{\mathsfit}{\encodingdefault}{\sfdefault}{m}{sl}
\SetMathAlphabet{\mathsfit}{bold}{\encodingdefault}{\sfdefault}{bx}{n}

\def\gE{{\mathcal{E}}}

\def\gN{{\mathcal{N}}}

\def\gS{{\mathcal{S}}}

\def\sR{{\mathbb{R}}}

\newcommand{\E}{\mathbb{E}}

\def\paperID{13063} % *** Enter the Paper ID here
\def\confName{CVPR}
\def\confYear{2024}

\title{Retinex-Diffusion: On Controlling Illumination Conditions in Diffusion Models via Retinex Theory}

\author{Xiaoyan Xing$^{1}$, Vincent Tao Hu$^{3}$\footnotemark[1], Jan Hendrik Metzen$^{2}$ \\ Konrad Groh$^{2}$, Sezer Karaoglu$^{1}$, Theo Gevers$^{1}$\\ \\{1. UvA-Bosch Delta Lab 2. Bosch Center for Artificial Intelligence 3. LMU Munich}}

\begin{document}

\twocolumn[{%
\renewcommand\twocolumn[1][]{#1}%
\maketitle
\begin{center}
\resizebox{\textwidth}{!}{
\includegraphics[width=\textwidth]{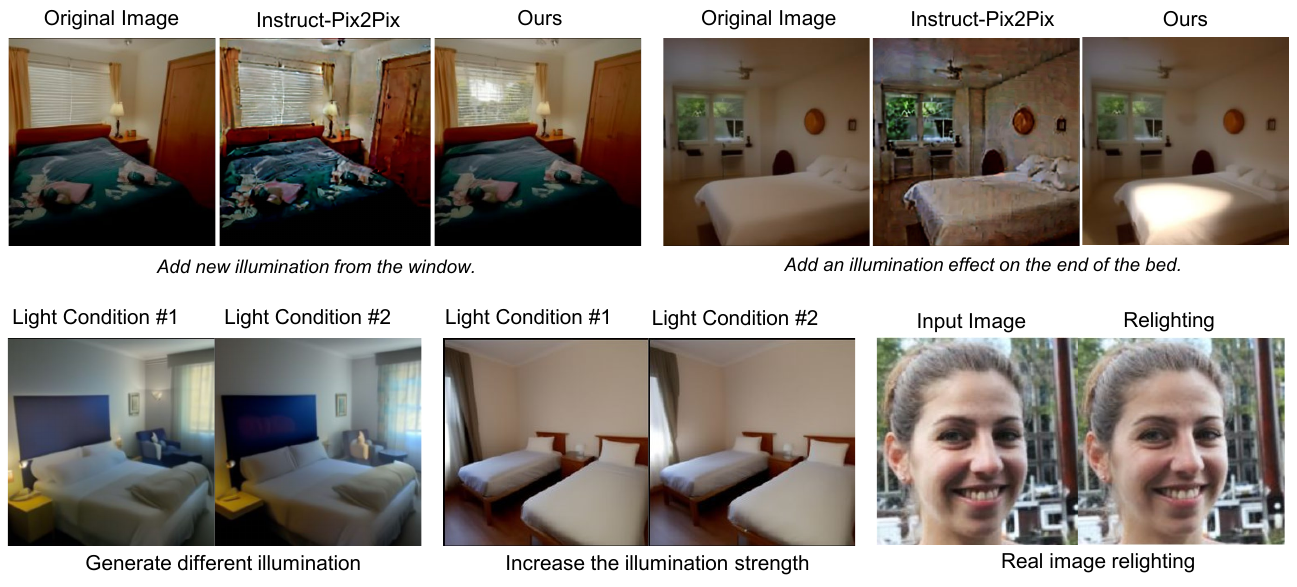}}

\vspace{0.3mm}
\captionof{figure}{Motivated by the recent state-of-the-art image manipulation method \cite{brooks2023instructpix2pix} can not handle the low-level manipulation, such as relighting the scene, or creating illumination effect (top row). We present physics-guided and training-free diffusion for controlling illumination conditions in images.  In image synthesis, our method generates photo-realistic illumination conditions under the proper illumination property guidance. Our model is also able to perform illumination editing of the original images, such as adding new illumination to images or face relighting (bottom row). Our approach is training-free and easily integrated with most pixel-based diffusion models, enhancing their illumination control capabilities efficiently.}
\vspace{1mm}
\label{fig:teaser}
\end{center}%
}]
\maketitle
\renewcommand{\thefootnote}{\fnsymbol{footnote}}
\vspace*{-3ex}
\footnotetext{$^*$Work was done at University of Amsterdam.} 
\renewcommand{\thefootnote}{1}
\begin{abstract}
\vspace{-2ex}
This paper introduces a novel approach to illumination manipulation in diffusion models, addressing the gap in conditional image generation with a focus on lighting conditions. We conceptualize the diffusion model as a black-box image render and strategically decompose its energy function in alignment with the image formation model. Our method effectively separates and controls illumination-related properties during the generative process. It generates images with realistic illumination effects, including cast shadow, soft shadow, and inter-reflections. Remarkably, it achieves this without the necessity for learning intrinsic decomposition, finding directions in latent space, or undergoing additional training with new datasets.
\end{abstract}    
\section{Introduction}
%The Gap
Generative models have shown their ability to create images that closely resemble real ones. The advent of conditional diffusion models has further enhanced the generation of specific semantic content \cite{mokady2022null,hertz2022prompt,kim2022diffusionclip,kwon2023diffusion, couairon2022diffedit}, contents \cite{huberman2023edit,meng2021sdedit,brooks2023instructpix2pix,rombach2022high_latentdiffusion_ldm}, layout \cite{zhang2023adding_controlnet,sgdm}, etc. However, a notable limitation is their inability to control illuminations precisely in the generated images. On the other hand, physically-based rendering pipelines like Blender \cite{blender2018} can achieve high-fidelity illumination but are time-consuming and lack diversity. In this paper, we combine the generative capabilities of diffusion models with the guidance of physically-based models to create a novel approach for controlling illumination conditions in both generated and real images.

%What other did to control
Altering illumination conditions is essential for various computer vision and computer graphics tasks. Typically, this involves decomposing and recomposing the intrinsic components of a scene \cite{li2022physically, zhu2022irisformer, li2021openrooms, ponglertnapakorn2023difareli, futschik2023controllable,luo2020niid}. Recently, the advancement of the generative models provide a way to solve such tasks end-to-end, e.g., finding the corresponding direction in the Style-GAN's latent space \cite{bhattad2022stylitgan, bhattad2023make}. However, these methods often require extensive datasets containing various intrinsic labels such as surface normals, depth, light sources, or the direction searching. In contrast, we propose a more straightforward approach to illumination condition control by harnessing the power of diffusion models guided by physical principles  (Table \ref{tab:advantage}). 

In this paper, we propose harnessing the generative process of the diffusion model as a self-contained rendering pipeline, enabling the manipulation of illumination conditions through physics-driven guidance. This approach allows us to circumvent the intricate tasks of decomposing and recomposing scene intrinsic components. Consequently, it can be seamlessly applied to diverse datasets and lighting scenarios without the need for additional training, tuning, or extra data labels.

In summary, the contributions are as follows:
\begin{compactitem}
\item We regard the diffusion model as a black-box image render and decompose its energy function according to the image formation model.

\item We introduce illumination-controllable image synthesis and geometry-preserving real image relighting, both guided by physics-based principles.

\item Our work provides photorealistic control of illumination in both generated and real images, delivering results that are on par with, and in some aspects surpass, models tailored to specific data domains.

\item Our approach is entirely devoid of training requirements and does not rely on any CGI techniques to attain controllable illumination conditions.
\end{compactitem}

\section{Related work}
\noindent \textbf{Conditional Diffusion Models.} Diffusion models perform well in various tasks due to their controllability~\cite{zhang2023adding_controlnet}. This includes image content~\cite{meng2021sdedit}, image layout~\cite{rombach2022high_latentdiffusion_ldm}, audio content~\cite{liu2023audioldm}, and human motion generation~\cite{tevet2022human_mdm}.
However, current diffusion models often struggle with fine-grained control over illumination as illustrated in Figure \ref{fig:teaser}. Our work leverages the knowledge of illumination to formulate flexible energy functions.
Classifier-guidance~\cite{dhariwal2021diffusion} demonstrates that diffusion models can be guided by pretraining a noisy-data based classifier. Classifier-free guidance~\cite{ho2021classifier} further eliminates the need for extra pretraining by randomly dropping out the guidance signal during training. Self-guided diffusion models~\cite{sgdm} even remove the annotation process altogether. More recently, ControlNet \citep{zhang2023adding_controlnet} and its variants has been used to address the illumination condition in diffusion models \citep{kocsis2024lightit,zeng2024dilightnet}, yet those methods usually require training Controlnet on extra dataset, resulting in a complex and costly implementation.

Our method is complementary to all these approaches. To fully leverage the potential of pretrained diffusion models, we propose an additional energy function based on illumination theory to guide the generation of the diffusion models.

% We summarize the illumination condition control methods, with respect to their technical approaches.
% % , the discussion on recent intrinsic decomposition methods and diffusion models are also provided. 

\noindent \textbf{Diffusion Based Image manipulation.} A substantial body of literature is dedicated to image manipulation and editing using diffusion models \cite{kim2022diffusionclip, brooks2023instructpix2pix, hertz2022prompt, meng2021sdedit, mokady2022null}. While these methods harness the remarkable generative capabilities of diffusion models to modify high to mid-level image attributes such as semantics~\cite{ho2020denoising, kwon2023diffusion} and layouts~\cite{rombach2022high_latentdiffusion_ldm, zhang2023adding_controlnet}, the control of lower-level features, especially illumination, has been unexplored. To this end, we introduce a novel illumination manipulation method tailored for conditional image generation based on illumination. Our approach involves decomposing the illumination-related properties during the generative process and applying targeted guidance based on these properties.

Notably, our training-free method can be easily integrated into major diffusion models, enhancing their ability to adjust illumination conditions on-the-fly. This contrasts with \cite{ponglertnapakorn2023difareli}, which is limited to human faces and relies on trained and pre-trained parameters.

\noindent \textbf{Intrinsic Image Decomposition.} Intrinsic image decomposition methods \cite{li2022physically,zhu2022irisformer,luo2020niid,barron2015shape,janner2017self, cgintrinsic, zhang2021nerfactor, ye2023intrinsicnerf,das2022pie, baslamisli2018cnn,baslamisli2021shadingnet,baslamisli2018joint} aim to decompose an image into illumination-relevant and illumination-irrelevant parts. Recent diffusion-based methods \cite{zeng2024rgb,kocsis2024iid} have shown impressive decomposition results, indicating that diffusion models have an awareness of intrinsic. 

Our method builds on one of the earliest vision techniques, retinex theory \citep{land1971lightness}, for solving intrinsic decomposition. It extracts the illumination property of the image and approximates it as the illumination-irrelevant component of the image at a very low cost. We employ the simple decomposition results as hints to guide the pre-trained diffusion models to generate images with designated illumination.
% \noindent{\textbf{Multiple steps methods.}} A common approach to control the illumination condition \cite{li2022physically,zhu2022irisformer,luo2020niid,barron2015shape,janner2017self, cgintrinsic, zhang2021nerfactor, ye2023intrinsicnerf} is to decompose an input image into its intrinsic components, e.g., reflectance, shading, surface normal, etc, and then recompose the image back with the modified intrinsic. 

% \noindent{\textbf{End-to-end methods.}}

\begin{table}[t]
  \centering
  \resizebox{\linewidth}{7ex}{
    \begin{tabular}{l|cccc}
    \toprule
          & \multicolumn{1}{c}{Neural} & \multicolumn{1}{c}{Directional } &  \multicolumn{1}{c}{DiFaReli} & Ours\\
          &  rendering\cite{zhu2022irisformer,li2022physically}     &  search \cite{bhattad2023make, bhattad2022stylitgan}    &\cite{ponglertnapakorn2023difareli}  &\\
    \midrule
    Neural backbone & CNN, Transformer    &  Style-GAN        & Diffusion & Diffusion\\
    label-free & light mask, depth    &        \checkmark   & light source, cam paras &  \checkmark\\
    End to End &     $\times$       &\checkmark   &  $\times$ &\checkmark\\
    Different types of data &   $\times$      &    $\times$   & $\times$  &\checkmark\\
    Diverse ouputs &   $\times$     &    Some of them   & $\times$  &\checkmark\\
    \end{tabular}}%
    \caption{Comparison with previous illumination control methods. }
  \label{tab:advantage}%
  \vspace{-4ex}
\end{table}%

% \noindent{\textbf{End-to-end methods.}}
% One can achieve such control by first estimate its intrinsic components using a neural network \cite{luo2020niid, zhu2022irisformer,li2022physically,janner2017self,cgintrinsic} and then 
% \subsection{Image }
\section{Preliminaries}

To illustrate the challenges in modifying illumination attributes within diffusion models, we begin by comparing the image synthesis process of diffusion models with the  image formation of real-world images.

\subsection{Real-world Image Formation}

% Given an RGB image $\textbf{I}$, the image formation can be described as follow:
% \begin{equation}
%     \textbf{I} = \int_{\omega_i \in \Omega+} \int_{\lambda} (N, \omega_i) E(\omega_i,\lambda)S(\omega_i,\omega_o,\lambda)R_{RGB}(\lambda)d\lambda d\omega,
% \end{equation}
The image formation can be modelled by~\cite{shafer1985using}:
% \begin{equation}
% I = I_b + I_s.
% \end{equation}
% The pixel at each point of the surface, measured over the visible spectrum $\omega$, is calculated by:
\begin{equation}
\begin{aligned}
    I = m(\mathbf{n},\mathbf{l})\int_\omega{f_c(\lambda)e(\lambda)\rho(\lambda)d\lambda}
    % +\\m_s(\mathbf{n},\mathbf{s},\mathbf{v})\int_\omega{f_c(\lambda)e(\lambda)\rho_s(\lambda)d\lambda},
\end{aligned}
\label{Equ: light_int}
\end{equation}
where $\mathbf{n}$ represents the surface normal vector, and $\mathbf{l}$ denotes the direction of the light source. The term $m$ represents the geometric relationship in the interaction. Additionally, $\lambda$ indicates the wavelength, $f_c(\lambda)$ the camera spectral sensitivity, $e(\lambda)$ the spectral power distribution of the illuminant, $\rho$ the surface reflectance.

Assuming the sensor response is linear and the wavelength of visible spectrum is narrow ($\lambda_I$), Equation \eqref{Equ: light_int} is simplified as:
\begin{equation}
% \begin{aligned}
    I = m(\mathbf{n},\mathbf{l}){e(\lambda_I)\rho(\lambda_I)}.
% \end{aligned}
\label{Equ: light_sum}
\end{equation}
Then, at point $\mathbf{x}$ on the surface, the decomposition of $I(\mathbf{x})$ can be approximated by:
\begin{equation}
    I(\mathbf{x}) = R(\mathbf{x})\odot{S(\mathbf{x})},
\label{Equ: light_diffuse}
\end{equation}

% \begin{equation}
%     A(\mathbf{x}) = R(\mathbf{x})\odot{I(\mathbf{x})},
% \label{Equ: light_diffuse}
% \end{equation}
where $\odot$ represents the Hadamard product, $R(\mathbf{x})$ is the reflectance, which is illumination-invariant, and $S(\mathbf{x})$ is the shading, which is illumination-variant. 
By manipulating shading, one can control the illumination properties in a physics-based manner.
% Furthermore, the model can be extend for a non-canonical light source (differ from Eq. \ref{Equ: light_sum}) as follows:
% \begin{equation}
%     I_b(\mathbf{x}) = R(\mathbf{x})\otimes{S(\mathbf{x})}\otimes{E(\mathbf{x})},
% \end{equation}
% where $E(\mathbf{x})$ describes the color of the light source at the point $\mathbf{x}$. Additionally, we can extend the Eq. \ref{Equ: light_diffuse} to non-diffuse reflection by adding the specular surface term $H(\mathbf{x})$:
% \begin{equation}
%     I(\mathbf{x}) = R(\mathbf{x})\otimes{S(\mathbf{x})}+H(\mathbf{x}),
% \end{equation}
% and for the non-canonical light source by:
% \begin{equation}
%     I(\mathbf{x}) = R(\mathbf{x})\otimes{S(\mathbf{x})}\otimes{E(\mathbf{x})} + H(\mathbf{x})\otimes{E(\mathbf{x})}.
% \end{equation}
% Finally, at the each point $\mathbf{x}$ of the surface, the illumination property $I_L$ of image $I$ is proportional to its shading, specular and color of the light source:
% \begin{equation}
%     I_L(\mathbf{x}) \propto[S(\mathbf{x})+H(\mathbf{x})]\otimes{E(\mathbf{x})}. 
% \end{equation}
% By manipulating shading, specular reflection, and the color of the light source, one can control the illumination properties in a physically-based manner.
\subsection{Diffusion Image Generation}

Diffusion models~\cite{sohl2015deep,ho2020denoising,song2021scorebased_sde} gradually perturb data using a forward diffusion process and then reverse the process to reconstruct the original data, as explained in previous studies~\cite{song2021maximum,ho2020denoising,dhariwal2021diffusion,bao2021analytic}. Let $q(\vx_0)$ denote the unknown data distribution in $\sR^D$. The forward diffusion process, indexed by time $t$ as  $\{\vx_t\}_{t \in [0,T]}$, can be succinctly represented by the following forward Stochastic Differential Equation (SDE):

\begin{align}
\label{eq:forward-sde}
    d\vx = \vf(\vx,t)dt + g(t)d\vw,
\end{align}
where $\vw \in \mathbb{R}^D$ is a standard Wiener process, $\vf(\cdot,t): \mathbb{R}^D \to \mathbb{R}^D$ is the drift coefficient and $g(t) \in \mathbb{R}$ is the diffusion coefficient. The $f(\vx,t)$ and $g(t)$ are related to the noise size and determine the perturbation kernel $q_{t|0}(\vx_t|\vx_0)$ from time $0$ to $t$. %In practice, the $f(\vy,t)$ is usually affine so that the the perturbation kernel is a linear Gaussian distribution and can be sampled in one step.

Let $q_t(\vx)$ be the marginal distribution of the SDE at time $t$ in Equation~\eqref{eq:forward-sde}. Its time reversal can be described by another SDE~\cite{song2021scorebased_sde}:
% The corresponding reverse-time diffusion models is give by the following SDE:
\begin{comment}
\begin{align}
\label{eq:reverse-sde}
    \mathrm{d} \vy = [\vf(\vy,t) - g(t)^2 \nabla_\vy \log q_t(\vy) ]\mathrm{d}t + g(t) \mathrm{d}\overline{\vw},
\end{align}
\end{comment}

\begin{align}
\label{eq:reverse-sde-score}
    \mathrm{d} \vx = [\vf(\vx,t) - g(t)^2 \vs(\vx, t) ]\mathrm{d}t + g(t) \mathrm{d}\overline{\vw},
\end{align}
where $\overline{\vw}$ is a reverse-time standard Wiener process with $\mathrm{d} t$ as an infinitesimal negative timestep, and $\vs(\vx, t)=\nabla_\vx \log q_t(\vx)$ represents the score. The score, similar to energy, allows us to introduce an additional energy function $\gE(\cdot,\cdot,\cdot)$ into the reverse SDE process for the specific guidance. We will further discuss the design of the energy function for illumination-based guidance in later sections.

\section{Method}
\begin{figure}
    \centering
    \includegraphics[width=0.5\textwidth]{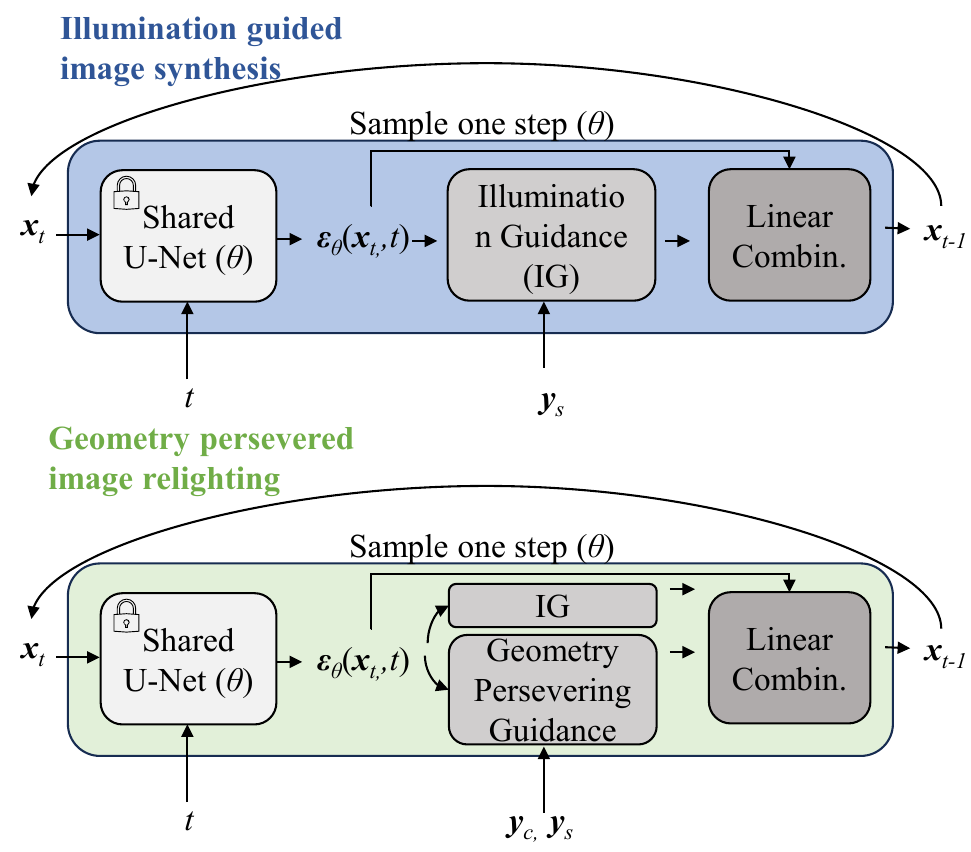}
    \caption{Overall diagram of the proposed illumination control diffusion method. Top, the illumination guidance image generation; Bottom, image relighting. $x_t$ represents the image at time step $t$, $\epsilon_\theta$ is the pre-trained U-Net \cite{unet}. $y_s$ and $y_c$ are prompts for illumination guidance and the geometry persevering guidance.}
    \label{fig: Diagram}
\end{figure}

As illustrated in Figure \ref{fig: Diagram}, our method performs two tasks: 1) it can control the illumination conditions of the generated image, and 2) it can apply new lighting conditions on real images. 
To accomplish these task, we first reformulate the energy function in the diffusion process. Then, we introduce the illumination guidance in image synthesis. Finally, we propose geometry preserved relighting for real images. Notably, this pipeline requires no further training, nor extra data labels or CGI techniques.

% \subsection{Intrinsic Aware Score Matching}
% We reconsider the reversed SDE processing of diffusion image generation through the lens of image formation. 
% As mentioned in the preliminary, an image can be decomposed into illumination-relevant $S$ and illumination-irrelevant $R$ parts. Motivated by this, we decompose the energy function $\gE(\vx, t)$ as the sum of two log potential functions~\cite{bishop:2006:PRML}: 
% \begin{equation}
%     \gE(\vx, t)  
%      =  \lambda_S\gE_{S}(\vx, t) +\lambda_R\gE_{R}(\vx, t) ,
% \end{equation}
% where $\lambda_S$ and $\lambda_R$ are weights for the illumination-relevant and illumination-irrelevant components, respectively. Hence, by manipulating the illumination-relevant energy function $\gE_{S}$, one could realize the illumination control. 

% Given an illumination prompt as target $\vy_s$, the reversed SDE processing of diffusion the generative process can be expressed as:
% \begin{equation}
%     \gE(\vy_s,\vx, t)  
%      =  \lambda_S\gE_{S}(\vy_s,\vx, t) +\lambda_R\gE_{R}(\vx, t) .
% \end{equation}
% Then the score function is:
% \begin{equation}
%     \nabla_{vx_s}\log{p_\theta}(\vx|\vy_s) = \nabla_{vx_s}\log{p_\theta}(\vx_s) + \lambda_S\cdot\nabla\log{{p_\theta}({\vy_s|\vx_s})}
% \end{equation}

\subsection{Illumination Energy Decomposition}
\label{Sec:illumination-aware}
We reconsider the reversed SDE processing of diffusion image generation based on the image formation model. As mentioned in the preliminary, an image can be decomposed into illumination-variant $I$ and illumination-invariant $R$ parts. Therefore, we decompose the energy function  $\gE(\vy, \vx, t)$ by the sum of two log potential functions~\cite{bishop:2006:PRML}: 
\begin{equation}
\begin{aligned}
     \label{eq:decompose energy}
     \gE(\vy, \vx, t)  
     &=  \lambda_I\gE_{I}(\vy, \vx, t) + \lambda_R\gE_{R}(\vy, \vx, t)  \\
     &= \lambda_I \E_{q_{t|0} (\vx_t|\vx)} \gS_{I}(\vy, \vx_t, t) \\
     &+ \lambda_R \E_{q_{t|0} (\vx_t|\vx)} \gS_{R}(\vy, \vx_t, t) ,
 \end{aligned}
\end{equation}
 where $\vy$ is our target guidance for energy, $\gE_{I}(\cdot, \cdot, \cdot):   \mathbb{R}^D \times \mathbb{R}^D  \times \mathbb{R} \rightarrow \mathbb{R}$ and $\gE_{R}(\cdot, \cdot, \cdot):  \mathbb{R}^D \times \mathbb{R}^D \times \mathbb{R} \rightarrow \mathbb{R}$ are the log potential functions, $\vx_t$ is the perturbed source image in the forward SDE, 
$q_{t|0} (\cdot|\cdot)$ is the perturbation kernel from time $0$ to time $t$ in the forward SDE, $\gS_{R}(\cdot, \cdot, \cdot):  \mathbb{R}^D \times \mathbb{R}^D \times \mathbb{R} \rightarrow \mathbb{R}$ and $\gS_{I}(\cdot, \cdot, \cdot): \mathbb{R}^D \times \mathbb{R}^D\times \mathbb{R} \rightarrow \mathbb{R}$ are functions measuring similarity between the target guidance and perturbed source image. $\lambda_R\in \mathbb{R}{>0}$ and $\lambda_S \in \mathbb{R}{>0}$ are weighting hyper-parameters for illumination-variant and illumination-invariant components, respectively. %Manipulating $\lambda_I$ controls illumination, while $\lambda_R$ controls non-illumination aspects.

In the reverse process, given by Equation~(\ref{eq:reverse-sde-score}), and adopting a step size of $h$, the iteration rule from $s$ to $t=s-h$ is as follows when applying our new illumination-based energy function:
\begin{equation}
\begin{aligned}
\label{eq:sampling-egsde}
\vx_t &= \vx_s - [\vf(\vx,s) - g(s)^2 (\vs(\vx_s,s) - \nabla_\vx \gE(\vy, \vx,s))] h \\ &+ g(s) \sqrt{h} \vz,
\end{aligned}
\end{equation}
where $\vz \sim \gN(\vzero, \mI)$.The expectation in $\gE(\vy, \vx,s)$ is estimated by the Monte Carlo method of a single sample for efficiency.  In the following section, we will introduce the construction of the energy function based on related illumination theories.

%For brevity, we present the general sampling procedure of our method in Algorithm~{\ref{alg:general}}. 

% \subsection{Illumination Property Extraction}

%Firstly, we consider the illumination-energy for sampled image generation, we introduce the illumination energy, and 

\subsection{Illumination-Energy Guided Generation}
\label{Sec: generate illumination}
%The illumination-guided generation, building upon intrinsic-aware score matching, depends on two modules: 1) illumination property extraction, which involves extracting illumination-related information from the diffusion sampling, and 2) proper guidance targeting, aiming at instructing the diffusion model to generate the desired lighting conditions.

% Why do it?
% Extracting the illumination property during the image generation processing is essential for better illumination control since it is objective to the illumination difference. 
% What others did and why it isn't work?
\paragraph{Prompt for Illumination Energy} Illumination is related to the brightness of a pixel. \cite{ponglertnapakorn2023difareli} shows that by matching the mean value difference between the reversed sampling and generated sampling, the consistency of global brightness of human portrait can be realized. This can be explained by Equation \eqref{Equ: light_diffuse}, where the reflectance $R$ of the human portrait can be approximated as a constant. Therefore, the intensity of pixel is directly reflecting illumination $S$. However, this simplification falls short when dealing with more complex surface reflectance. Therefore, an additional step of illumination property extraction is required. 

% $f_r(x_t)$

We modify the multi-scale retinex model \cite{rahman1996multi} to extract illumination properties.
The illumination is computed from the denosied image $x_t$ at the candidate step $t$. First $x_t \in \mathbb{R}^{H\times{W}\times3}$ is filtered by 2D Gaussians at multiple scales, to capture the image's varying level of geometry and illumination. Subsequently, these blurred images are combined using either uniform or user-defined weights. The processing is formed as:
\begin{equation}
    f_s(m_t,n_t) = \sum_{r,g,b}\sum_{k=1}^{N} w_k \cdot \left( \frac{1}{2 \pi \sigma_k^2} e^{-\frac{i^2 + j^2}{2 \sigma_k^2}} * x_t(m,n) \right),
\end{equation}
where $f_s(m_t,n_t)$ is the estimated lighting at pixel $(m,n)$ at given timestep $t$, $\{i,j\}$ represents the position of the Gaussian center, $\sigma_k$ is the standard deviation for the $k^{th}$ scale, and $*$ donates the convolution.  

\paragraph{Illumination Energy Guidance}
% In terms of prompting the illumination condition to the pretrained diffusion model, one could modify the latent or the diffusion model itself \cite{ponglertnapakorn2023difareli}. However, both options require the data pair of illumination labels and corresponding images, which is inefficient, and can not be scaled to other dataset. 
% In terms of prompting the illumination condition to the pretrained diffusion model, one could encode the illumination condition to the 
We model the target illumination condition by parameterizing the illumination map as a composition of $N$ 2D Gaussian functions $G(\cdot)$ as:
\begin{equation}
\label{Eq: illum_target}
    y_s = \sum_{i=1}^{N} \alpha_i G(\mu_i, \Sigma_i),
\end{equation}
where its mean $\mu_i$ represents the position of the light source (for the visible light source) or the location of the brightest parts of the image (for the invisible light source), and its covariance matrix ($\Sigma_i$) describes the spread and directionality of the light. $\alpha_i$ is the weight of corresponding Gaussian, which is subject to $\Sigma{\alpha_i}=1$.

%\paragraph{Illumination Energy Guidance in Sampling}

After the illumination property $f_s(.)$ and the illumination prompt $y_s$ are computed, their difference can be quantified using any differentiable similarity metric. In practice, the pixel-wise mean square error is used to calculate this difference, which is defined as:
\begin{equation}
    \label{eq:energy_iii}
    \gS_{I}(y_s, \vx, t) = \sum_{x \in \vx} ||f_s(x_t)-y_{s}||_2.
\end{equation}
% \begin{equation}
%     \label{eq:energy_iii}
%     \gS_{I}(\vy, \vx, t) = \sum_{x \in \vx} ||\hat{S}(x_t)-\vy||_2. w.r.t  \vy = S_{target}
% \end{equation}

% \tao{todo, add t}
% %Similarly to the classifier gradient guidance in \cite{dhariwal2021diffusion_beat}, we evaluate the anti-gradient $-\nabla_{S_t}\mathcal{L}$\tao{TODO} to bring the illumination guidance to the diffusion process.

% \tao{TODO} In terms of the guided step, in contrast to the common strategy, which is to apply the guidance in the beginning steps, we apply our guidance from the midden step to several steps ahead of the final step. We argue that, by applying such a guidance strategy, our method is able to obtain the proper illumination guidance results without limiting the diversity of the content of the generative results. (Details in ablation study.)

\subsection{Illumination-Energy on Real-world Image}
\label{Sce: edit}
The illumination-based guidance generation can be extended to modify or add lighting conditions in real-world images, as shown in Figure~\ref{fig: Diagram}. This involves two key steps: firstly, an inverse processing stage to convert the input image into a latent code, and secondly, a reflectance-based correction to maintain the original image geometry while altering the lighting. This approach enables flexible lighting manipulation while preserving the original image's core structure and details. 

\paragraph{Real-image inversion}

%We follow the deterministic reverse DDIM process as introduced by \cite{dhariwal2021diffusion,song2020denoising_ddim,kim2022diffusionclip} to invert $x_0$ to latent space, then we .

% \begin{align}
%     \frac{ \vx_{t - \Delta t} }{ \sqrt{ \alpha_{ t - \Delta t} } } = \frac{ \vx_{t} }{ \sqrt{ \alpha_{t} } } + \left( \sqrt{ \frac{ 1 - \alpha_{ t - \Delta t} }{ \alpha_{ t - \Delta t} } } - \sqrt{ \frac{ 1 - \alpha_{ t } }{ \alpha_{ t } } } \right) \epsilon^{(t)}_{\theta}(\vx_{t})
% \end{align}
% where $\epsilon^{(t)}_{\theta}(\vx_{t})$ is the noise prediction, and $\alpha_t$ is a predefined noise schedule from \emph{variance-preserving} SDE~\cite{ho2020denoising} or \emph{variance-exploding} SDE~\cite{song2019generative}. For other mathematical details of diffusion models please refer to the supplemental. 
 Given an input image $x_0\in\mathbb{R}^{H\times{W}\times3}$, we utilize  the inverse process of DDIM~\cite{song2020denoising_ddim} converts it to a latent $x_z\in\mathbb{R}^{H\times{W}\times3}$ at the final forward DDIM step $z$. Hence, given $x_z$ as input, the illumination-guided generative process is able to change the light condition of the real image. Yet, the edited image may not have the exact same geometry, since the guidance on shading may also disturb the geometry. 

\paragraph{Geometry persevering guidance} Another key contribution of our method is the reflectance-based correction for preserving the geometry of the edited image. 
% As the reformed energy function in Equation \eqref{eq:decompose energy}, 
Similar to the condition on the illumination property, the geometry property can be regarded as part of illumination-invariant component ${R}$. And it can be represented as the surface reflectance difference. We apply a gradient-based method to extract such difference by calculating the cross color ratios (CCR) \cite{gevers1999color} of the surface. 

Given an image $x_0\in\mathbb{R}^{H\times{W}\times3}$ to be edited and two neighbouring pixels $p_1$, $p_2$. The CCR's are calculated as:
% ($M_{RG}$) of the $(R,G)$ channel pairs is 
\begin{equation}
\label{eq:CCR}
    M_{RG}=\frac{R_{p_1}G_{p_2}}{R_{p_2}G_{p_1}},
    M_{RB}=\frac{R_{p_1}B_{p_2}}{R_{p_2}B_{p_1}},
    M_{GB}=\frac{G_{p_1}B_{p_2}}{G_{p_2}B_{p_1}},
\end{equation}
where, $M_{RG},M_{RB}$ and $ M_{GB}$ are the CCR for the color channel pairs $(R,G)$, $(R,B)$ and $(G,B)$, respectively. 
% can be replaced by $(R,B)$ and $(G,B)$ to obtain the CCR for other color pairs. 
To simplify the writing, we use $M_{RG}$ to represent the CCR for all channel pairs, in the following statements. 
Given the image formation in Equation \eqref{Equ: light_sum}, and the illumination condition are close in a small neighborhood \cite{das2022pie}, we obtain:
% Taking the logarithm on both side of the equation, we get:
% \begin{equation}
%     \log{M_{RG}} = \log{R_{p_1}}+\log{G_{p_2}}-\log{R_{p_2}}-\log{G_{p_1}}.
% \end{equation}
% Given the image formation in Eq. \ref{Equ: light_int}, at certain point, we obtain:
% \begin{equation}
%     C_{p_1} = m_b(\mathbf{n},\mathbf{s}){e^{C_{p_1}}(\lambda)\rho_b^{C_{p_1}}(\lambda)}+\\m_s(\mathbf{n},\mathbf{s},\mathbf{v}){e^{C_{p_1}}(\lambda)\rho_s^{C_{p_1}}(\lambda)},
% \end{equation}
% where $C_{p_1}$ represents the color channel $C$ at pixel $p_1$, for an input image. Since the two neighbouring pixels $p_1$ and $P_2$, the same illumination condition can be assumed. Therefore, 
% \begin{equation}
%     e^{C_{p_1}} = e^{C_{p_2}}.
% \end{equation}
% Then, the Eq. can be reformed as:
\begin{equation}
\label{eq: log_ccr}
\begin{aligned}
    \log{M_{RG}} = \log(\rho^{R_{p_1}})+\log(\rho^{G_{p_2}})\\-\log(\rho^{R_{p_2}})-\log(\rho^{G_{p_1}}).
\end{aligned}
\end{equation}
Therefore, the CCR are illumination invariant and only depends on reflectance transitions. For more mathematical details, please refer to the supplementary material section. The final normalized target CCR matrix
$\textbf{C}_{x_0}\in\mathbb{R}^{H\times{W}\times3}$ is [$M_{RG},M_{RB},M_{GB}$]. The operation to compute the CCR matrix for a image is expressed as: $f_c(\cdot)$.

Similarly, the CCR matrix of the generated image at time step $t$ can be also calculated in the same manner, denoted by $\textbf{C}_{x_t} = f_c(x_t)$.

%\paragraph{Geometry preserved relighting}
In a manner similar to the illumination prompting, we define the geometry-related prompt as $y_c$. Hence, the discrepancy between the target CCR $y_c$ and the computed CCR $f_c(x_t)$ at sampling step $t$ is quantified as follows:

\begin{equation}
\label{eq:energy_rrr}
 \gS_{R}(y_c,\vx_t, t)  = \sum_{x \in \vx} ||f_c(x_t)-y_c)||_2.
\end{equation}
Then, the anti-gradient of the reflectance difference $-\nabla_{\textbf{C}_t}\mathcal{L}$ is sent to the diffusion process together with the anti-gradient of the illumination difference $-\nabla_{S_t}\mathcal{L}$, to achieve the geometry preserved relighting in the real image editing.

% With diffusion inversion, the our method can be extended to the real-world image editing. Gievn an image $x_0\in\mathbb{R}^{H\times{W}\times3}$, the DDIM inversion is able to reverse the image into the latent space. We can then apply the guided sampling on top of the space. Unlike the image synthesis which allows a diverse generalization in terms of the semantic information. The image editing requires a geometric preserve. To achieve that, we first 
\section{Experiments}

% For the image generated by ADM~\cite{dhariwal2021diffusion_beat} and DDIM~\cite{song2020denoising_ddim_ddim}, the image size is set to $256\times256$. For the image generated by EDM~\cite{karras2022elucidating}, the image size is $64\times64$. 

\paragraph{Pre-trained diffusion models.} Our approach leverages the generative capabilities of pre-trained diffusion models. A summary of the models used in our study, along with the datasets on which they are pre-trained, is as follows:
\begin{compactitem}
    \item \textbf{ADM} \cite{dhariwal2021diffusion}: pretrained on LSUN-Bedroom\cite{yu15lsun} for generating and editing the diverse illumination conditions of bedrooms.
    \item \textbf{EDM} \cite{karras2022elucidating}: pretrained on FFHQ dataset for generating diverse illumination conditions of human faces.
    \item \textbf{DDIM} \cite{song2020denoising_ddim}: pretrained on CelebA-HQ \cite{ffhq_karras2019style} for editing illumination conditions of human faces.
\end{compactitem}
\begin{figure*}
    \centering
    \includegraphics[width=\textwidth]{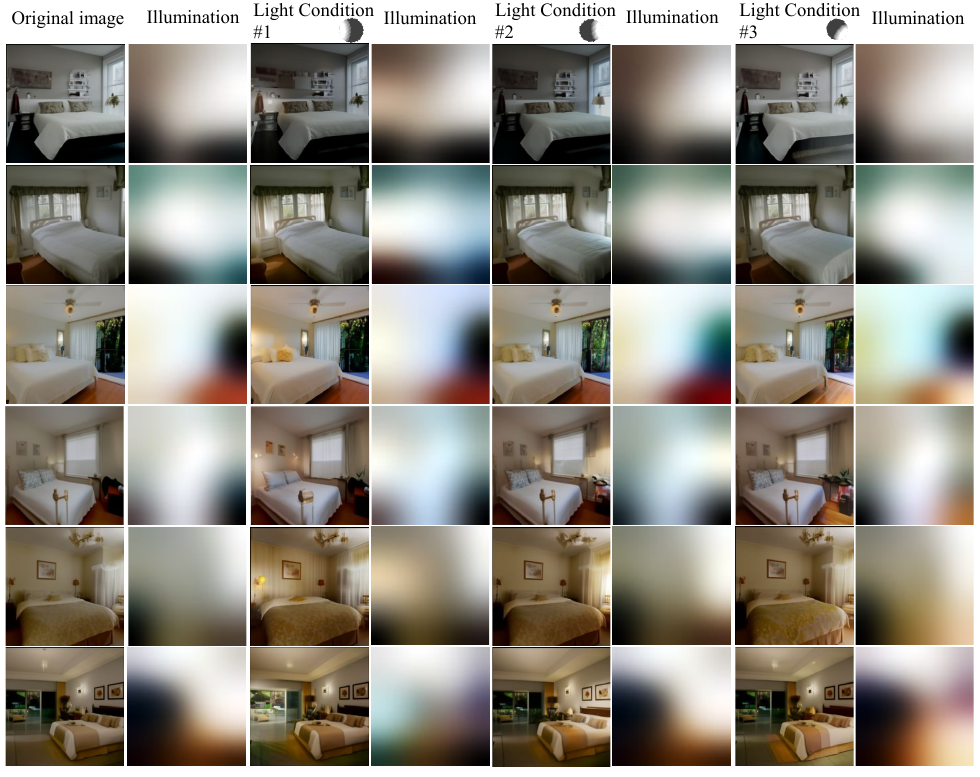}
    \vspace{-2ex}
    \caption{Illumination Property-Guided Image Generation: Each pair of columns displays the generated image alongside its corresponding illumination feature. The initial two columns show the original image without illumination guidance and its illumination feature. Subsequent columns illustrate images generated under various specific lighting conditions, with the illumination direction indicated by a sphere.}
    \label{fig:main-show}
\end{figure*}
\begin{figure*}
    \centering
    \includegraphics[width=0.9\textwidth]{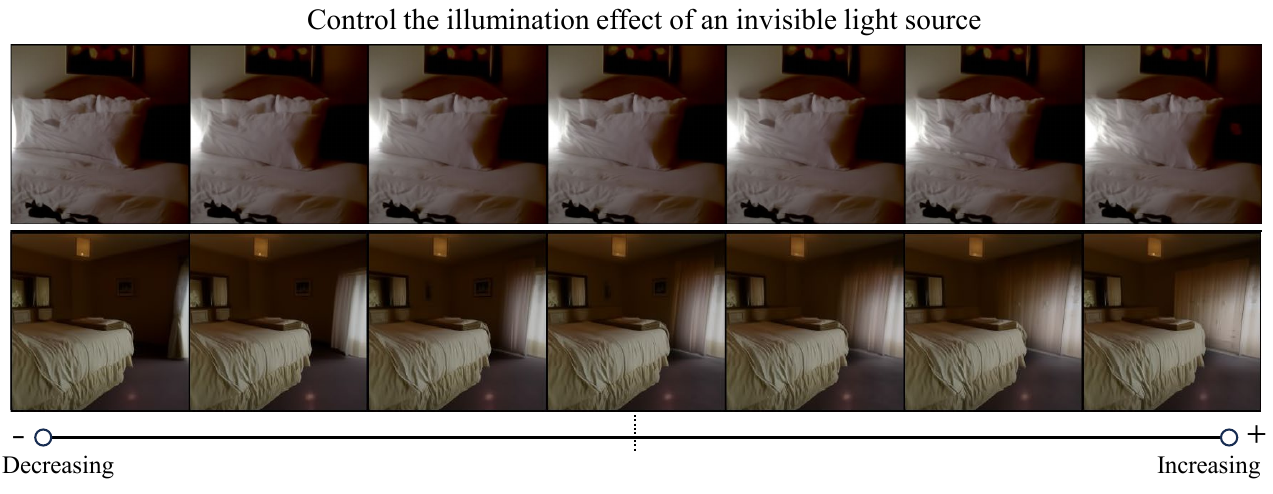}
    \caption{Illumination effect control of an invisible light source. Given the same illumination direction prompt, our method is able to generate multiple variants of illumination effect with respect to the strength of the light source.}
    \label{fig:main-control}
\end{figure*}
\subsection{Illumination Guidance in Image Synthesis}
First, we demonstrate the effectiveness of our method in generating photo-realistic images with different illumination variants, including different light direction, shadow, and strength of illumination.  Then, we evaluate the illumination guidance by comparing it with the human-instructed image to image generative model \cite{brooks2023instructpix2pix}.
In the case of the human-instructed model, we employ text prompts that specify fixed illumination conditions to generate images. In contrast, our method directly integrates illumination prompts to guide the generation process.

\begin{figure}
    \centering
    \includegraphics[width=0.8\linewidth]{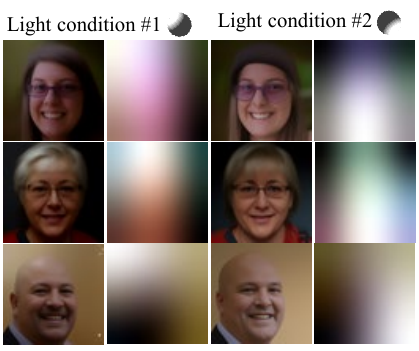}
    \caption{Illumination Property-Guided Image Generation (Using EDM \cite{karras2022elucidating}): Each pair of columns displays the generated image alongside its corresponding illumination feature. Illumination direction indicated by a sphere. }
    \label{fig:main-face}
    \vspace{-2ex}
\end{figure}

Figure \ref{fig:main-show} and Figure \ref{fig:main-face} illustrate the images generated under varying illumination condition guidance. The associated illumination features are displayed alongside the images. The results showcase the photo-realistic generation of illumination conditions, such as turning on a lamp or increasing outdoor illumination, and so on.

Figure \ref{fig:main-control} highlights our method's application in controlling the illumination effect of a new light source. By using consistent illumination direction prompts, it illustrates the ability to produce a range of illumination effects. 
These variations correspond to the intensity of the light source, illustrating the flexibility of our method in adjusting light intensity within a scene.

\begin{figure}
    \centering
    \includegraphics[width=0.9\linewidth]{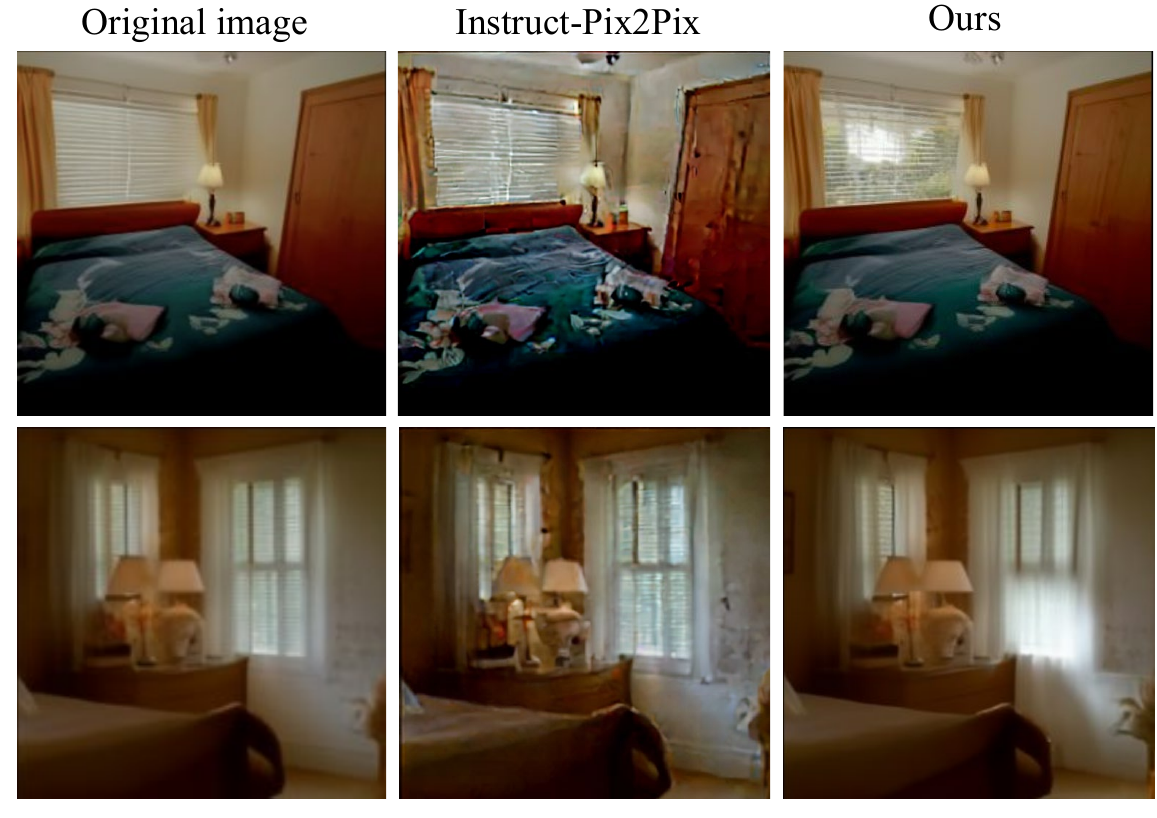}
    \caption{Visual comparison with state-of-the-art image to image diffusion model \cite{brooks2023instructpix2pix} in the context of generating new lighting conditions.}
    \label{fig:main-ldm}
\end{figure}

Figure \ref{fig:main-ldm} visually demonstrates the results of increasing the illumination coming from the window. These results can be compared with those obtained using human-guided image-to-image generative methods, Instruct-Pix2Pix \cite{brooks2023instructpix2pix}, which struggles on generating the image with the prompted illumination, our method generates vivid illumination with our illumination guided generation. The hyper-parameters of \cite{brooks2023instructpix2pix} are carefully selected to obtain the stable results. 

% To quantitatively assess the effectiveness of this illumination guidance, we introduce the Illumination Map Difference (IMD) metric. The IMD metric, denoted as $\mathcal{L}_{IMD}$, quantifies the performance of the guidance by measuring the discrepancy between the generated illumination and the target. It is defined as follows:

% \begin{equation}
% \mathcal{L}_{IMD} = \frac{1}{N}\sum{||f_s({x_0})-y_s||_2},
% \end{equation}

Table \ref{Tab:illum_MSE} presents the quantitative evaluation of our method for generating images under specified illumination conditions. We use MSE as the metric to measure the discrepancy between the illumination extracted from the generated images and the given illumination prompt. Across three distinct illumination conditions, our method consistently achieves a better MSE score compared to the baseline method without guidance and \cite{brooks2023instructpix2pix} with our carefully designed prompt, underscoring the effectiveness of ours.
\begin{table}
    \centering
    \resizebox{\linewidth}{!}{
    \begin{tabular}{l|ccc}
    \toprule
    % & MSE
         & Illumination & Illumination & Illumination  \\
         &  condition \#1 & condition \#2 & condition \#3 \\
    \midrule     
         Instruct-Pix2Pix \cite{brooks2023instructpix2pix}& 0.0896 &0.0835  & 0.0346\\
         ADM \cite{dhariwal2021diffusion}&0.0887   & 0.0823&0.0344  \\
         Ours (ADM+illum. guided) &\textbf{0.0757}  &\textbf{0.0706}  & \textbf{0.0311}  \\
    \bottomrule
    \end{tabular}}
    \caption{Quantitative results of the illumination map generated by Instruct-Pix2Pix\cite{brooks2023instructpix2pix}(w. text prompt guidance), ADM (w/o. illumination guidance) and ours (w. illumination guidance). }
    \label{Tab:illum_MSE}
\end{table}

\subsection{Illumination Editing in Real-world Image}
\begin{figure*}
    \centering
    \includegraphics[width=\linewidth]{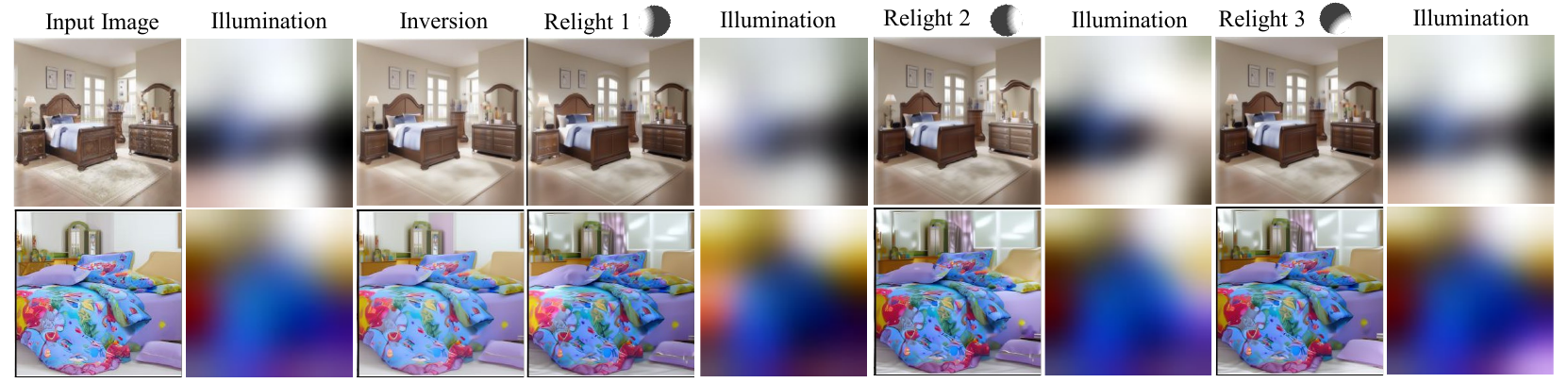}
    \caption{Geometry-Preserved Image Relighting: The sequence from left to right displays the original input image, the inverted image, and the image relighting results under three different lighting conditions (denoted as Relight 1, Relight 2, and Relight 3). This progression demonstrates the effectiveness of our method in maintaining geometric consistency while altering illumination conditions. Illumination direction indicated by a sphere.}
    \label{fig:main-relight}
\end{figure*}

We further evaluate our methods on illumination editing in synthetic and real-world images.
Figure \ref{fig:main-relight} demonstrates our method's capability in illumination editing on various images. The sequence comprises the original input image, its inversion, and the final relighted image, each accompanied by their corresponding illumination features. Remarkably, our method skillfully adjusts the illumination direction while maintaining the original geometry, showcasing its effectiveness even on colorful surfaces.

\begin{figure}
    \centering
    \includegraphics[width=\linewidth]{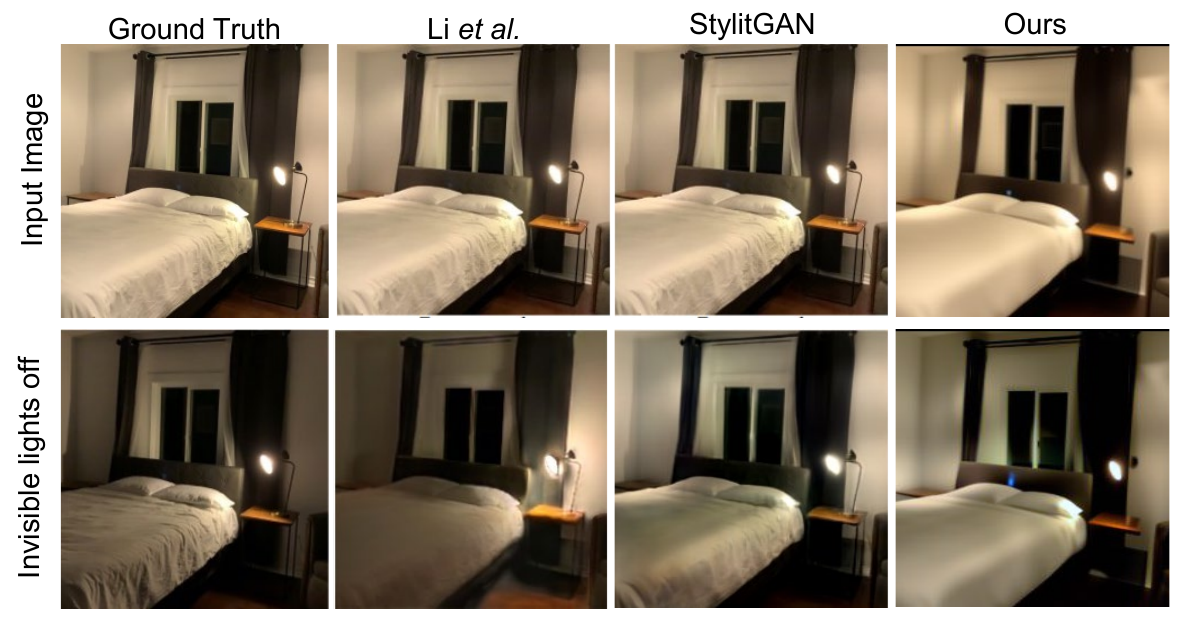}
    \caption{Real indoor image relighting comparison: The left column displays real images captured with invisible light turned on and off, illustrating the ground truth. The middle-left column showcases images relighted by the method proposed by Li et al. \cite{li2022physically}, which utilizes multiple sub-networks for intrinsic decomposition and generates new images based on these modified intrinsics. The middle-right column shows StylitGAN's \cite{bhattad2022stylitgan} relighting, applying directional search in the latent space. The right column presents our results using illumination-guided diffusion, notably achieving relighting without learned intrinsic decomposition, directional search, or CGI techniques.}
    \label{fig:real-relight-room}
    \vspace{-2ex}
\end{figure}

Figure \ref{fig:real-relight-room} shows real-world image relighting using our method, which operates without learned intrinsics \cite{li2022physically} or directional searches in latent space \cite{bhattad2022stylitgan}. The loss of high-frequency detail is attributed to the limitations of DDIM inversion, as our guidance does not significantly alter the geometry.

\begin{figure}
    \centering
    \includegraphics[width=\linewidth]{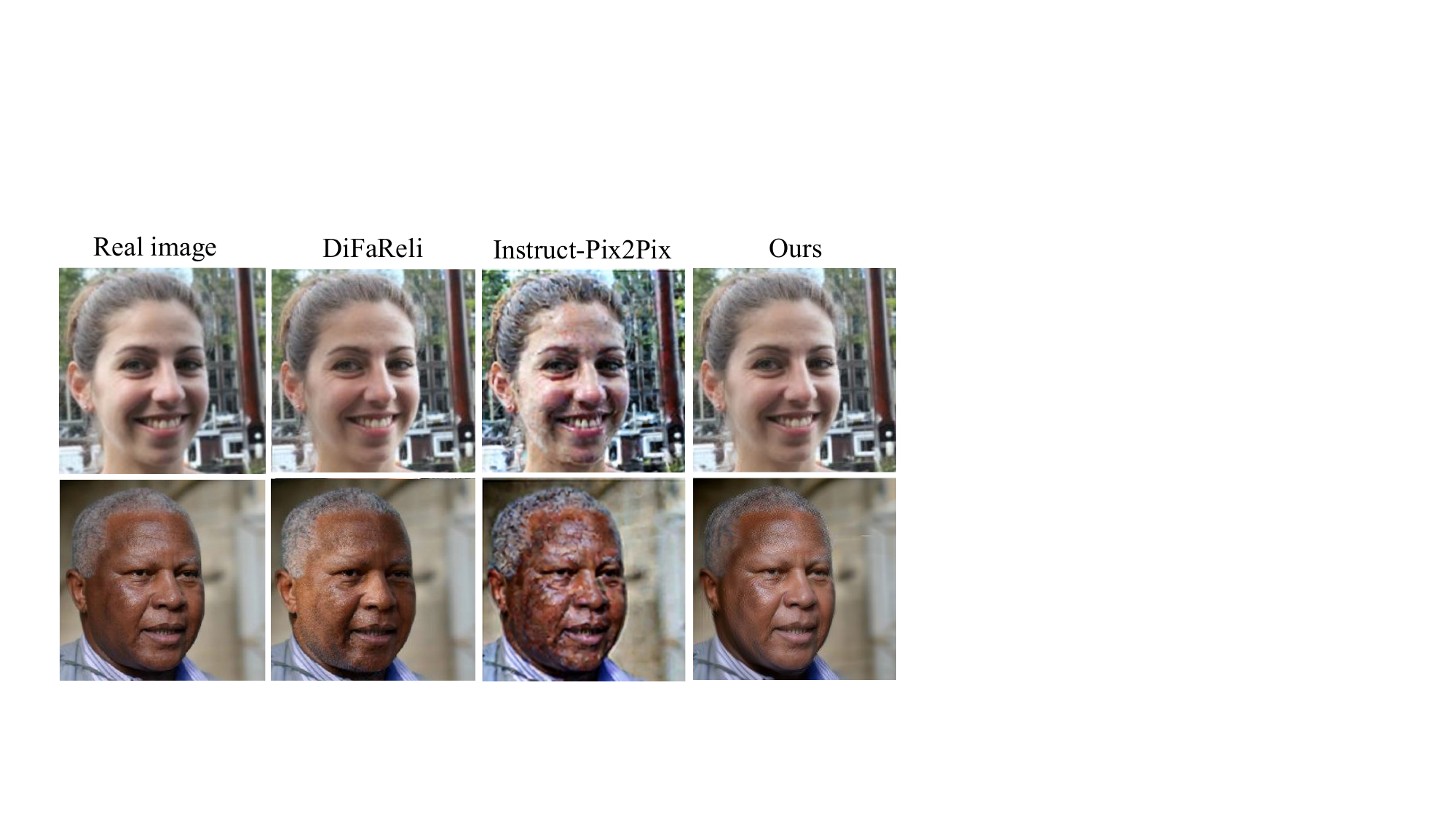}
    \caption{Real face image relighting comparison: The left column presents the real face image for relighting. Middle-left column, the lateset diffusion based face relighting method \cite{ponglertnapakorn2023difareli}, which is conditioned on multiple intrinsic and extrinsic components, and needs multiple pretrained methods to decompose those components. Middle-right column, human instructed image to image generative method \cite{brooks2023instructpix2pix}, fails badly on relighting. Right column, our method is able to relight the face without learned intrinsic decomposition, directional search, or CGI techniques. }
    \label{fig:real-relight-face}
\end{figure}

Figure \ref{fig:real-relight-face} showcases our real-face relighting results. Our method surpasses the performance of the state-of-the-art image-to-image manipulation method by \cite{brooks2023instructpix2pix} (we carefully designed their text prompt.). Additionally, our results are comparable to those obtained by the specialized face relighting method of \cite{ponglertnapakorn2023difareli}.

\begin{figure}
    \centering
    \includegraphics[width=0.8\linewidth]{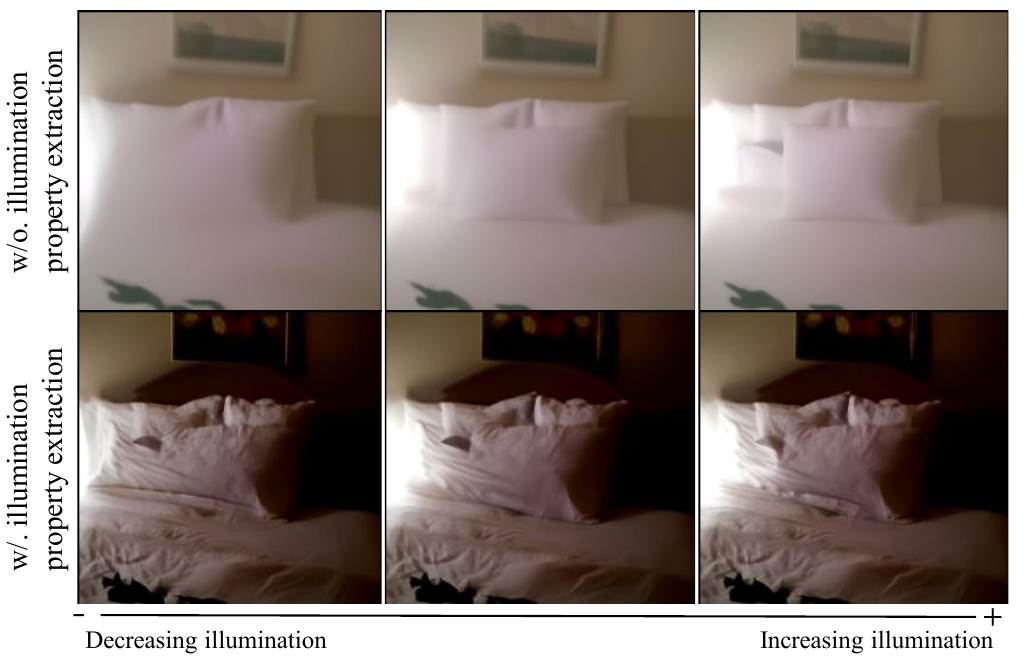}
    \caption{Ablation comparison of the illumination property extraction. Top row: illumination guided generation without extracting the illumination property. Bottom row: illumination guided generation with the extracted illumination property.}
    \label{fig:alb-illum_property}
    \vspace{-2ex}
\end{figure}
\subsection{Ablation Study \& Discussion}

\noindent{\textbf{Illumination property extraction.}} 
As introduced in Section \ref{Sec: generate illumination} , by extracting the illumination property, one can distinguish the illumination condition from the composited image generation. Here, we evaluate the illumination guidance without extracting the illumination property. As shown in Figure \ref{fig:alb-illum_property}, we guide the image synthesis process based on the same illumination prompts, while we gradually increase the lighting range of the light prompts, the content of the image guided by the non-illumination specific property shifts rapidly. Also, images are much blurry compared to the ones with extracting illumination properties. Such results clearly showcase the effectiveness of the illumination property extraction.

\begin{table}
    \centering
    \resizebox{\linewidth}{!}{
    \begin{tabular}{l|cccc}
    \toprule
         Sampler& Batch & Sampling & Batch Time (s) & Batch Time (s) \\
         &  Size & steps & w/o. guidance & w. guidance\\
    \midrule     
         ADM \cite{dhariwal2021diffusion}& 16 & 100 & 102.63 &110.85\\
         DDIM \cite{song2020denoising_ddim}& 8 & 40 & 11.59 &15.63\\
         EDM \cite{karras2022elucidating}& 64 & 15 &  4.17 &4.35\\
    \bottomrule
    \end{tabular}}
    \caption{Comparison of generation times (per batch) with and without illumination guidance using ADM \cite{dhariwal2021diffusion}, DDIM \cite{song2020denoising_ddim}, and EDM \cite{karras2022elucidating} on a Pytorch framework. Tests were conducted on an Nvidia RTX4090 GPU.}
    \label{tab:computational_cost}
\end{table}
\noindent{\textbf{Computational efficiency.}}
The computational cost analysis of our method (Table \ref{tab:computational_cost}), reveals minimal additional burden on the guided diffusion model. Specifically, the inference time for image generation/editing, using standard resolutions, batch sizes, and sampling steps, shows only a marginal increase. An increment of 8.2 seconds is shown for ADM \cite{dhariwal2021diffusion}, 0.17 seconds for EDM \cite{karras2022elucidating}, and 4.04 seconds for DDIM \cite{song2020denoising_ddim}, affirming the efficiency of ours.

\noindent{\textbf{Other intrinsic decomposition methods.}} Theoretically, the retinex model can be replaced by any other decomposition models, including  learning-based models. We employ the multiscale retinex model for following reason: 1) it is naturally differentiable, and can be directly used to calculate the score; 2) it is computational cheap, since it only contains few convolution operations, it will not significantly increase the sampling time. 
% \noindent{Illumination Property Guided Generation.} 
% \noindent{Illumination Property Guided Generation.} 
\section{Conclusion}

We proposed a novel and physics-based, training-free method for manipulating illumination in both diffusion-generated and real images. Our method realizes accurate illumination-conditioned image generation by reforming the energy function of diffusion models based on the image formation model. Our approach does not require any type of extra training or direction research. It is easy to be embedded into current diffusion models. By prompting the illumination-related feature, the diffusion model is able to generate/adjust illumination-related semantics for proper illumination conditions, such as turning on the lamp and opening the window. 

\noindent{\textbf{Limitation}.} Such illumination control is not always accurate, and it is aligned to the learned data distribution of  pretrained diffusion models. A deeper investigation on extracting intrinsics of the diffusion models is needed in the future.  
%Moreover, the performance of our method aligns to the generative ability of the corresponding diffusion models, artifacts, outliners are also expected for those out-of-distribution data. 

{
    \small
    \bibliographystyle{ieeenat_fullname}
    \bibliography{main}

\begin{thebibliography}{52}
\providecommand{\natexlab}[1]{#1}
\providecommand{\url}[1]{\texttt{#1}}
\expandafter\ifx\csname urlstyle\endcsname\relax
  \providecommand{\doi}[1]{doi: #1}\else
  \providecommand{\doi}{doi: \begingroup \urlstyle{rm}\Url}\fi

\bibitem[Bao et~al.(2021)Bao, Li, Zhu, and Zhang]{bao2021analytic}
Fan Bao, Chongxuan Li, Jun Zhu, and Bo Zhang.
\newblock Analytic-dpm: an analytic estimate of the optimal reverse variance in diffusion probabilistic models.
\newblock In \emph{International Conference on Learning Representations}, 2021.

\bibitem[Barron and Malik(2015)]{barron2015shape}
Jonathan~T Barron and Jitendra Malik.
\newblock Shape, illumination, and reflectance from shading.
\newblock \emph{IEEE TPAMI}, 37\penalty0 (8):\penalty0 1670--1687, 2015.

\bibitem[Baslamisli et~al.(2018{\natexlab{a}})Baslamisli, Groenestege, Das, Le, Karaoglu, and Gevers]{baslamisli2018joint}
Anil~S Baslamisli, Thomas~T Groenestege, Partha Das, Hoang-An Le, Sezer Karaoglu, and Theo Gevers.
\newblock Joint learning of intrinsic images and semantic segmentation.
\newblock In \emph{ECCV}, 2018{\natexlab{a}}.

\bibitem[Baslamisli et~al.(2018{\natexlab{b}})Baslamisli, Le, and Gevers]{baslamisli2018cnn}
Anil~S Baslamisli, Hoang-An Le, and Theo Gevers.
\newblock Cnn based learning using reflection and retinex models for intrinsic image decomposition.
\newblock In \emph{CVPR}, 2018{\natexlab{b}}.

\bibitem[Baslamisli et~al.(2021)Baslamisli, Das, Le, Karaoglu, and Gevers]{baslamisli2021shadingnet}
Anil~S Baslamisli, Partha Das, Hoang-An Le, Sezer Karaoglu, and Theo Gevers.
\newblock Shadingnet: Image intrinsics by fine-grained shading decomposition.
\newblock \emph{IJCV}, 129\penalty0 (8):\penalty0 2445--2473, 2021.

\bibitem[Bhattad and Forsyth(2022)]{bhattad2022stylitgan}
Anand Bhattad and David~A Forsyth.
\newblock Stylitgan: Prompting stylegan to produce new illumination conditions.
\newblock \emph{arXiv preprint arXiv:2205.10351}, 2022.

\bibitem[Bhattad et~al.(2023)Bhattad, Shah, Hoiem, and Forsyth]{bhattad2023make}
Anand Bhattad, Viraj Shah, Derek Hoiem, and DA Forsyth.
\newblock Make it so: Steering stylegan for any image inversion and editing.
\newblock \emph{arXiv preprint arXiv:2304.14403}, 2023.

\bibitem[Bishop(2006)]{bishop:2006:PRML}
Christopher~M. Bishop.
\newblock \emph{Pattern Recognition and Machine Learning}.
\newblock 2006.

\bibitem[Brooks et~al.(2023)Brooks, Holynski, and Efros]{brooks2023instructpix2pix}
Tim Brooks, Aleksander Holynski, and Alexei~A Efros.
\newblock Instructpix2pix: Learning to follow image editing instructions.
\newblock In \emph{CVPR}, pages 18392--18402, 2023.

\bibitem[Community(2018)]{blender2018}
Blender~Online Community.
\newblock \emph{Blender - a 3D modelling and rendering package}.
\newblock Blender Foundation, Stichting Blender Foundation, Amsterdam, 2018.

\bibitem[Couairon et~al.(2022)Couairon, Verbeek, Schwenk, and Cord]{couairon2022diffedit}
Guillaume Couairon, Jakob Verbeek, Holger Schwenk, and Matthieu Cord.
\newblock Diffedit: Diffusion-based semantic image editing with mask guidance.
\newblock In \emph{ARXIV}, 2022.

\bibitem[Das et~al.(2022)Das, Karaoglu, and Gevers]{das2022pie}
Partha Das, Sezer Karaoglu, and Theo Gevers.
\newblock Pie-net: Photometric invariant edge guided network for intrinsic image decomposition.
\newblock In \emph{CVPR}, 2022.

\bibitem[Dhariwal and Nichol(2021)]{dhariwal2021diffusion}
Prafulla Dhariwal and Alexander Nichol.
\newblock Diffusion models beat gans on image synthesis.
\newblock \emph{NeurIPS}, 34:\penalty0 8780--8794, 2021.

\bibitem[Futschik et~al.(2023)Futschik, Ritland, Vecore, Fanello, Orts-Escolano, Curless, S{\`y}kora, and Pandey]{futschik2023controllable}
David Futschik, Kelvin Ritland, James Vecore, Sean Fanello, Sergio Orts-Escolano, Brian Curless, Daniel S{\`y}kora, and Rohit Pandey.
\newblock Controllable light diffusion for portraits.
\newblock In \emph{CVPR}, pages 8412--8421, 2023.

\bibitem[Gevers and Smeulders(1999)]{gevers1999color}
Theo Gevers and Arnold~WM Smeulders.
\newblock Color-based object recognition.
\newblock \emph{Pattern recognition}, 32\penalty0 (3):\penalty0 453--464, 1999.

\bibitem[Hertz et~al.(2022)Hertz, Mokady, Tenenbaum, Aberman, Pritch, and Cohen-Or]{hertz2022prompt}
Amir Hertz, Ron Mokady, Jay Tenenbaum, Kfir Aberman, Yael Pritch, and Daniel Cohen-Or.
\newblock Prompt-to-prompt image editing with cross attention control.
\newblock \emph{arXiv preprint arXiv:2208.01626}, 2022.

\bibitem[Ho and Salimans(2021)]{ho2021classifier}
Jonathan Ho and Tim Salimans.
\newblock Classifier-free diffusion guidance.
\newblock In \emph{NeurIPS Workshop}, 2021.

\bibitem[Ho et~al.(2020)Ho, Jain, and Abbeel]{ho2020denoising}
Jonathan Ho, Ajay Jain, and Pieter Abbeel.
\newblock Denoising diffusion probabilistic models.
\newblock \emph{NeurIPS}, 2020.

\bibitem[Hu et~al.(2023)Hu, Zhang, Asano, Burghouts, and Snoek]{sgdm}
Vincent~Tao Hu, David~W Zhang, Yuki~M. Asano, Gertjan~J. Burghouts, and Cees G.~M. Snoek.
\newblock Self-guided diffusion models.
\newblock In \emph{CVPR}, 2023.

\bibitem[Huberman-Spiegelglas et~al.(2023)Huberman-Spiegelglas, Kulikov, and Michaeli]{huberman2023edit}
Inbar Huberman-Spiegelglas, Vladimir Kulikov, and Tomer Michaeli.
\newblock An edit friendly ddpm noise space: Inversion and manipulations.
\newblock \emph{arXiv}, 2023.

\bibitem[Janner et~al.(2017)Janner, Wu, Kulkarni, Yildirim, and Tenenbaum]{janner2017self}
Michael Janner, Jiajun Wu, Tejas~D Kulkarni, Ilker Yildirim, and Josh Tenenbaum.
\newblock Self-supervised intrinsic image decomposition.
\newblock In \emph{NIPS}, 2017.

\bibitem[Karras et~al.(2019)Karras, Laine, and Aila]{ffhq_karras2019style}
Tero Karras, Samuli Laine, and Timo Aila.
\newblock A style-based generator architecture for generative adversarial networks.
\newblock In \emph{CVPR}, 2019.

\bibitem[Karras et~al.(2022)Karras, Aittala, Aila, and Laine]{karras2022elucidating}
Tero Karras, Miika Aittala, Timo Aila, and Samuli Laine.
\newblock Elucidating the design space of diffusion-based generative models.
\newblock In \emph{NeurIPS}, 2022.

\bibitem[Kim et~al.(2022)Kim, Kwon, and Ye]{kim2022diffusionclip}
Gwanghyun Kim, Taesung Kwon, and Jong~Chul Ye.
\newblock Diffusionclip: Text-guided diffusion models for robust image manipulation.
\newblock In \emph{CVPR}, 2022.

\bibitem[Kocsis et~al.(2024{\natexlab{a}})Kocsis, Philip, Sunkavalli, Nie{\ss}ner, and Hold-Geoffroy]{kocsis2024lightit}
Peter Kocsis, Julien Philip, Kalyan Sunkavalli, Matthias Nie{\ss}ner, and Yannick Hold-Geoffroy.
\newblock Lightit: Illumination modeling and control for diffusion models.
\newblock In \emph{CVPR}, 2024{\natexlab{a}}.

\bibitem[Kocsis et~al.(2024{\natexlab{b}})Kocsis, Sitzmann, and Nie{\ss}ner]{kocsis2024iid}
Peter Kocsis, Vincent Sitzmann, and Matthias Nie{\ss}ner.
\newblock Intrinsic image diffusion for single-view material estimation.
\newblock In \emph{CVPR}, 2024{\natexlab{b}}.

\bibitem[Kwon et~al.(2023)Kwon, Jeong, and Uh]{kwon2023diffusion}
Mingi Kwon, Jaeseok Jeong, and Youngjung Uh.
\newblock Diffusion models already have a semantic latent space.
\newblock 2023.

\bibitem[Land and McCann(1971)]{land1971lightness}
Edwin~H Land and John~J McCann.
\newblock Lightness and retinex theory.
\newblock \emph{Josa}, 61\penalty0 (1):\penalty0 1--11, 1971.

\bibitem[Li and Snavely(2018)]{cgintrinsic}
Zhengqi Li and Noah Snavely.
\newblock Cgintrinsics: Better intrinsic image decomposition through physically-based rendering.
\newblock In \emph{ECCV}, 2018.

\bibitem[Li et~al.(2021)Li, Yu, Sang, Wang, Song, Liu, Yeh, Zhu, Gundavarapu, Shi, et~al.]{li2021openrooms}
Zhengqin Li, Ting-Wei Yu, Shen Sang, Sarah Wang, Meng Song, Yuhan Liu, Yu-Ying Yeh, Rui Zhu, Nitesh Gundavarapu, Jia Shi, et~al.
\newblock Openrooms: An open framework for photorealistic indoor scene datasets.
\newblock In \emph{CVPR}, pages 7190--7199, 2021.

\bibitem[Li et~al.(2022)Li, Shi, Bi, Zhu, Sunkavalli, Ha{\v{s}}an, Xu, Ramamoorthi, and Chandraker]{li2022physically}
Zhengqin Li, Jia Shi, Sai Bi, Rui Zhu, Kalyan Sunkavalli, Milo{\v{s}} Ha{\v{s}}an, Zexiang Xu, Ravi Ramamoorthi, and Manmohan Chandraker.
\newblock Physically-based editing of indoor scene lighting from a single image.
\newblock In \emph{ECCV}, pages 555--572. Springer, 2022.

\bibitem[Liu et~al.(2023)Liu, Chen, Yuan, Mei, Liu, Mandic, Wang, and Plumbley]{liu2023audioldm}
Haohe Liu, Zehua Chen, Yi Yuan, Xinhao Mei, Xubo Liu, Danilo Mandic, Wenwu Wang, and Mark~D Plumbley.
\newblock Audioldm: Text-to-audio generation with latent diffusion models.
\newblock \emph{arXiv preprint arXiv:2301.12503}, 2023.

\bibitem[Luo et~al.(2020)Luo, Huang, Li, Zhou, Zhang, and Bao]{luo2020niid}
Jundan Luo, Zhaoyang Huang, Yijin Li, Xiaowei Zhou, Guofeng Zhang, and Hujun Bao.
\newblock Niid-net: adapting surface normal knowledge for intrinsic image decomposition in indoor scenes.
\newblock \emph{IEEE TVCG}, 26\penalty0 (12):\penalty0 3434--3445, 2020.

\bibitem[Meng et~al.(2021)Meng, He, Song, Song, Wu, Zhu, and Ermon]{meng2021sdedit}
Chenlin Meng, Yutong He, Yang Song, Jiaming Song, Jiajun Wu, Jun-Yan Zhu, and Stefano Ermon.
\newblock Sdedit: Guided image synthesis and editing with stochastic differential equations.
\newblock In \emph{ICLR}, 2021.

\bibitem[Mokady et~al.(2022)Mokady, Hertz, Aberman, Pritch, and Cohen-Or]{mokady2022null}
Ron Mokady, Amir Hertz, Kfir Aberman, Yael Pritch, and Daniel Cohen-Or.
\newblock Null-text inversion for editing real images using guided diffusion models.
\newblock \emph{arXiv}, 2022.

\bibitem[Ponglertnapakorn et~al.(2023)Ponglertnapakorn, Tritrong, and Suwajanakorn]{ponglertnapakorn2023difareli}
Puntawat Ponglertnapakorn, Nontawat Tritrong, and Supasorn Suwajanakorn.
\newblock Difareli: Diffusion face relighting.
\newblock \emph{arXiv preprint arXiv:2304.09479}, 2023.

\bibitem[Rahman et~al.(1996)Rahman, Jobson, and Woodell]{rahman1996multi}
Zia-ur Rahman, Daniel~J Jobson, and Glenn~A Woodell.
\newblock Multi-scale retinex for color image enhancement.
\newblock In \emph{ICIP}, pages 1003--1006. IEEE, 1996.

\bibitem[Rombach et~al.(2022)Rombach, Blattmann, Lorenz, Esser, and Ommer]{rombach2022high_latentdiffusion_ldm}
Robin Rombach, Andreas Blattmann, Dominik Lorenz, Patrick Esser, and Bj{\"o}rn Ommer.
\newblock High-resolution image synthesis with latent diffusion models.
\newblock In \emph{CVPR}, 2022.

\bibitem[Ronneberger et~al.(2015)Ronneberger, Fischer, and Brox]{unet}
Olaf Ronneberger, Philipp Fischer, and Thomas Brox.
\newblock U-net: Convolutional networks for biomedical image segmentation.
\newblock In \emph{MICCAI}, 2015.

\bibitem[Shafer(1985)]{shafer1985using}
Steven~A Shafer.
\newblock Using color to separate reflection components.
\newblock \emph{Color Research \& Application}, 10\penalty0 (4):\penalty0 210--218, 1985.

\bibitem[Sohl-Dickstein et~al.(2015)Sohl-Dickstein, Weiss, Maheswaranathan, and Ganguli]{sohl2015deep}
Jascha Sohl-Dickstein, Eric Weiss, Niru Maheswaranathan, and Surya Ganguli.
\newblock Deep unsupervised learning using nonequilibrium thermodynamics.
\newblock In \emph{ICML}, 2015.

\bibitem[Song et~al.(2021{\natexlab{a}})Song, Meng, and Ermon]{song2020denoising_ddim}
Jiaming Song, Chenlin Meng, and Stefano Ermon.
\newblock Denoising diffusion implicit models.
\newblock In \emph{ICLR}, 2021{\natexlab{a}}.

\bibitem[Song et~al.(2021{\natexlab{b}})Song, Durkan, Murray, and Ermon]{song2021maximum}
Yang Song, Conor Durkan, Iain Murray, and Stefano Ermon.
\newblock Maximum likelihood training of score-based diffusion models.
\newblock \emph{NeurIPS}, 34:\penalty0 1415--1428, 2021{\natexlab{b}}.

\bibitem[Song et~al.(2021{\natexlab{c}})Song, Sohl-Dickstein, Kingma, Kumar, Ermon, and Poole]{song2021scorebased_sde}
Yang Song, Jascha Sohl-Dickstein, Diederik~P Kingma, Abhishek Kumar, Stefano Ermon, and Ben Poole.
\newblock Score-based generative modeling through stochastic differential equations.
\newblock In \emph{ICLR}, 2021{\natexlab{c}}.

\bibitem[Tevet et~al.(2022)Tevet, Raab, Gordon, Shafir, Cohen-Or, and Bermano]{tevet2022human_mdm}
Guy Tevet, Sigal Raab, Brian Gordon, Yonatan Shafir, Daniel Cohen-Or, and Amit~H Bermano.
\newblock Human motion diffusion model.
\newblock \emph{arXiv preprint arXiv:2209.14916}, 2022.

\bibitem[Ye et~al.(2023)Ye, Chen, Bao, Bao, Pollefeys, Cui, and Zhang]{ye2023intrinsicnerf}
Weicai Ye, Shuo Chen, Chong Bao, Hujun Bao, Marc Pollefeys, Zhaopeng Cui, and Guofeng Zhang.
\newblock Intrinsicnerf: Learning intrinsic neural radiance fields for editable novel view synthesis.
\newblock In \emph{Proceedings of the IEEE/CVF International Conference on Computer Vision}, pages 339--351, 2023.

\bibitem[Yu et~al.(2015)Yu, Zhang, Song, Seff, and Xiao]{yu15lsun}
Fisher Yu, Yinda Zhang, Shuran Song, Ari Seff, and Jianxiong Xiao.
\newblock Lsun: Construction of a large-scale image dataset using deep learning with humans in the loop.
\newblock In \emph{ARXIV}, 2015.

\bibitem[Zeng et~al.(2024{\natexlab{a}})Zeng, Dong, Peers, Kong, Wu, and Tong]{zeng2024dilightnet}
Chong Zeng, Yue Dong, Pieter Peers, Youkang Kong, Hongzhi Wu, and Xin Tong.
\newblock Dilightnet: Fine-grained lighting control for diffusion-based image generation.
\newblock In \emph{ACM SIGGRAPH 2024 Conference Proceedings}, 2024{\natexlab{a}}.

\bibitem[Zeng et~al.(2024{\natexlab{b}})Zeng, Deschaintre, Georgiev, Hold-Geoffroy, Hu, Luan, Yan, and Ha{\v{s}}an]{zeng2024rgb}
Zheng Zeng, Valentin Deschaintre, Iliyan Georgiev, Yannick Hold-Geoffroy, Yiwei Hu, Fujun Luan, Ling-Qi Yan, and Milo{\v{s}} Ha{\v{s}}an.
\newblock Rgb $\leftrightarrow$ x: Image decomposition and synthesis using material-and lighting-aware diffusion models.
\newblock \emph{arXiv preprint arXiv:2405.00666}, 2024{\natexlab{b}}.

\bibitem[Zhang et~al.(2023)Zhang, Rao, and Agrawala]{zhang2023adding_controlnet}
Lvmin Zhang, Anyi Rao, and Maneesh Agrawala.
\newblock Adding conditional control to text-to-image diffusion models.
\newblock In \emph{Proceedings of the IEEE/CVF International Conference on Computer Vision}, pages 3836--3847, 2023.

\bibitem[Zhang et~al.(2021)Zhang, Srinivasan, Deng, Debevec, Freeman, and Barron]{zhang2021nerfactor}
Xiuming Zhang, Pratul~P Srinivasan, Boyang Deng, Paul Debevec, William~T Freeman, and Jonathan~T Barron.
\newblock Nerfactor: Neural factorization of shape and reflectance under an unknown illumination.
\newblock \emph{ACM TOG}, 40\penalty0 (6):\penalty0 1--18, 2021.

\bibitem[Zhu et~al.(2022)Zhu, Li, Matai, Porikli, and Chandraker]{zhu2022irisformer}
Rui Zhu, Zhengqin Li, Janarbek Matai, Fatih Porikli, and Manmohan Chandraker.
\newblock Irisformer: Dense vision transformers for single-image inverse rendering in indoor scenes.
\newblock In \emph{CVPR}, pages 2822--2831, 2022.

\end{thebibliography}
}

\clearpage
\setcounter{page}{1}
\maketitlesupplementary
\appendix
\section{Cross Color Ratios}
Section \ref{Sce: edit} introduced one can use cross color ratios to represent part of the illumination-invariant components. Here we provide the detail to prove the CCR is illumination-invariant.
Taking the logarithm on both side of the Equation \eqref{eq:CCR}, we get:
\begin{equation}
    \log{M_{RG}} = \log{R_{p_1}}+\log{G_{p_2}}-\log{R_{p_2}}-\log{G_{p_1}}.
\end{equation}
Given the image formation in Equation \eqref{Equ: light_int} and \cite{baslamisli2018cnn}, at certain point, we obtain:
\begin{equation}
    C_{p_1} = m_b(\mathbf{n},\mathbf{s}){e^{C_{p_1}}(\lambda)\rho_b^{C_{p_1}}(\lambda)}+\\m_s(\mathbf{n},\mathbf{s},\mathbf{v}){e^{C_{p_1}}(\lambda)\rho_s^{C_{p_1}}(\lambda)},
\end{equation}
where $C_{p_1}$ represents the color channel $C$ at pixel $p_1$, for an input image. $\rho_b$ and $\rho_s$ represent the body diffuse refletance and specular refletance respectively. Since the two neighbouring pixels $p_1$ and $p_2$, the same illumination condition can be assumed. Therefore, 
\begin{equation}
    e^{C_{p_1}} = e^{C_{p_2}}.
\end{equation}

Then, Equation \eqref{eq: log_ccr} can be derived. It is important to note that the Cross-Color Ratio (CCR) is solely influenced by the surface reflectance. Consequently, the CCR remains invariant to changes in illumination.

\end{document}